\documentclass{article} 
\usepackage{iclr2026_conference,times}

\usepackage{amsmath,amsfonts,bm}

\def\eqref#1{equation~\ref{#1}}

\def\1{\bm{1}}

\DeclareMathAlphabet{\mathsfit}{\encodingdefault}{\sfdefault}{m}{sl}
\SetMathAlphabet{\mathsfit}{bold}{\encodingdefault}{\sfdefault}{bx}{n}

\usepackage[hidelinks]{hyperref}
\usepackage{url}
\usepackage{amsmath,amssymb}  
\usepackage{booktabs}         
\usepackage{graphicx}  
\usepackage{subcaption}  
\usepackage{algorithm}
\usepackage{algpseudocode}

\title{TangramSR: Can Vision-Language Models Reason in Continuous Geometric Space? }

\author{Yikun Zong \thanks{Corresponding author} \\
Department of Engineering\\
University of Cambridge\\
Cambridge, CB3 0DG, United Kingdom \\
\texttt{\{yz977\}@cam.ac.uk} \\
\And
Cheston Tan \\
Centre for Frontier AI Research \\
A*STAR \\
138632, Singapore \\
\texttt{\{cheston-tan\}@a-star.edu.sg} \\
}

\iclrfinalcopy 
\begin{document}

\maketitle

\begin{abstract}
Humans excel at spatial reasoning tasks like Tangram puzzle assembly through cognitive processes involving mental rotation, iterative refinement, and visual feedback. Inspired by how humans solve Tangram puzzles through trial-and-error, observation, and correction, we design a framework that models these human cognitive mechanisms. However, comprehensive experiments across five representative Vision--Language Models (VLMs) reveal \textbf{systematic failures in continuous geometric reasoning}: average IoU of only 0.41 on single-piece tasks, dropping to 0.23 on two-piece composition, far below human performance where children can complete Tangram tasks successfully. This paper addresses a fundamental challenge in \textbf{self-improving AI}: can models iteratively refine their predictions \emph{at test time} without parameter updates? We introduce a \textbf{test-time self-refinement framework} that combines in-context learning (ICL) with reward-guided feedback loops, inspired by human cognitive processes. Our \textbf{training-free verifier--refiner agent} applies \textbf{recursive refinement loops} that iteratively self-refine predictions based on geometric consistency feedback, achieving IoU improvements from 0.63 to 0.932 on medium-triangle cases \emph{without any model retraining}. This demonstrates that incorporating human-inspired iterative refinement mechanisms through ICL and reward loops can substantially enhance geometric reasoning in VLMs, moving self-improving AI from promise to practice in continuous spatial domains. Our work is available at this anonymous link https://anonymous.4open.science/r/TangramVLM-F582/.
\end{abstract}

\section{Introduction}

When arranging pieces of a Tangram puzzle, even a small positional or angular deviation can disrupt the target configuration. 
The task requires reasoning in \emph{continuous geometric space}, where success depends on accurately aligning orientations, adjusting fine-scale positions, and composing parts into a coherent whole~\citep{shepard1971mental}.
Despite their impressive performance on vision and language benchmarks~\citep{radford2021learning}, current Vision–Language Models (VLMs) still struggle in such settings.

\textbf{Human cognition as inspiration.}
Humans excel at spatial reasoning tasks like Tangram puzzle assembly through cognitive processes involving mental rotation, iterative refinement, and visual feedback~\citep{shepard1971mental,bohning1997using}. Despite their success on discrete benchmarks~\citep{radford2021learning}, current Vision--Language Models (VLMs) struggle in continuous geometric reasoning, revealing a fundamental gap between human cognitive capabilities and AI reasoning. Building on cognitive science research showing that humans solve spatial tasks through iterative refinement and feedback-driven self-correction~\citep{shepard1971mental,bohning1997using}, we design a benchmark where Tangram pieces must be precisely arranged to cover a target silhouette. Humans solve spatial reasoning tasks through a functional cognitive process: trial and error, observation, correction. This involves explicit decomposition (position, angle, size), mental simulation through mental rotation, and feedback-driven correction. Inspired by these human cognitive mechanisms, we design a framework that models how humans iteratively refine spatial estimates through visual feedback, incorporating the 'mental rotation' and 'iterative refinement' processes described by Shepard \& Metzler~\citep{shepard1971mental} into an AI reasoning pipeline.

\textbf{VLM failures in continuous space.}
Vision Language Models (VLMs) have achieved remarkable success in discrete reasoning tasks, from mathematical problem-solving~\citep{cobbe2021training} to code generation. However, their performance in \emph{continuous geometric reasoning}, tasks requiring precise spatial alignment with metric precision, remains largely unexplored. This gap is critical: real-world applications such as robotic manipulation, puzzle assembly, and spatial planning demand accurate coordinate predictions in continuous space, where small errors can lead to catastrophic failures. Unlike discrete reasoning where approximate answers may suffice, continuous geometric tasks require exact spatial relationships: positions must align within pixel-level precision, angles must match within degrees, and scales must preserve relative proportions. Tangram puzzle assembly, where small positional/angular deviations disrupt configurations, provides an ideal testbed for evaluating continuous geometric reasoning~\citep{shepard1971mental,bohning1997using,yamada2017comparative}. 
\textbf{Surprisingly, despite success in discrete reasoning}~\citep{cobbe2021training}, experiments across five VLMs (Qwen~\citep{bai2023qwen}, GPT-4o~\citep{hurst2024gpt,islam2025gpt}, LLaMA~\citep{gao2023llama}, Gemini, Claude) show \textbf{systematic failures}: only 0.41 IoU on single-piece tasks and 0.23 on two-piece composition, significantly low compared to human performance~\citep{bohning1997using}.

\textbf{Test-time self-improvement as a path forward.}
This paper addresses a fundamental challenge in \textbf{self-improving AI}: how can models iteratively refine their predictions \emph{at test time} without parameter updates? Traditional approaches to improving VLM performance rely on retraining or fine-tuning, requiring computational resources and training data that may not always be available. 
However, recent advances in \emph{self-improving AI} have demonstrated that models can enhance their capabilities \emph{at test time} through iterative refinement, feedback loops, and reward-guided corrections. 
In continuous geometric reasoning, small errors in position, angle, or scale can compound to produce large failures, yet these errors are often correctable through feedback-driven adjustments. 
This suggests a natural role for \textbf{test-time self-refinement}: enabling models to iteratively improve their geometric predictions based on geometric consistency feedback, without requiring parameter updates or retraining.
\textbf{While model parameters remain unchanged, changes occur in the ICL context and prompt content} as the refinement loop iteratively updates the in-context examples and feedback signals based on previous iterations, enhancing the model's geometric reasoning capabilities (tools and skills) through feedback-driven refinement at test time.

Our framework establishes a \textbf{test-time self-improvement pipeline} for VLMs operating in continuous geometric space, with applications in \textbf{scientific discovery} and robotics.  
First, it isolates fundamental geometric skills, estimating position, angle, and size using mathematically grounded metrics (\(\ell_2\) distance and IoU).  
Then, it escalates to two-piece arrangement, a compositional task that probes whether models can reason jointly about multiple parts.  
Through this lens, we uncover \textbf{systematic failure modes} across diverse models, Qwen~\citep{bai2023qwen}, GPT-4o~\citep{hurst2024gpt,islam2025gpt}, LLaMA~\citep{gao2023llama}, Gemini, and Claude, showing that even frontier systems perform badly when geometry becomes continuous, achieving only 0.41 IoU on single-piece tasks and 0.23 on two-piece composition, far below human performance~\citep{bohning1997using}.  
\textbf{ICL and reward loop as a solution.}
\textbf{Crucially}, we demonstrate that combining \textbf{in-context learning (ICL)} with \textbf{reward-guided feedback loops}---inspired by how humans use prior examples and receive visual feedback---can iteratively correct these errors, achieving substantial IoU improvements from 0.63 to 0.932 without any model training. While symbolic approaches such as AlphaGeometry~\citep{chervonyi2025gold} have reached Olympiad-level theorem proving, they operate in a discrete symbolic space of axioms and proofs.  
Our work, by contrast, situates reasoning directly in the continuous geometric world, where correctness is measured not by logic alone but by spatial alignment. We find that the gap between discrete success and continuous failure reveals a fundamental limitation of current VLMs: they can recognize geometric concepts, but struggle to compute and reason in a continuous, quantitative manner.

\paragraph{Contributions}
(1) \textbf{Human-inspired framework}: We introduce a framework inspired by how humans solve Tangram puzzles through cognitive processes (mental rotation, iterative refinement, visual feedback), modeling human trial-and-error correction mechanisms in an AI reasoning pipeline. 
(2) \textbf{Systematic evaluation revealing VLM failures}: We conduct a comprehensive evaluation across leading VLMs and find consistent and severe geometric failures in continuous space: \textbf{average IoU on single-piece tasks is only 0.41}, while \textbf{two-piece composition drops to 0.23}, far below human performance, revealing that even frontier systems cannot maintain geometric consistency in continuous space. 
(3) \textbf{ICL and reward loop solution}: We propose a lightweight, \textbf{recursive self-improvement framework} that combines in-context learning (ICL) with reward-guided refinement loops---inspired by how humans use prior examples and receive visual feedback---to progressively enhance geometric alignment \emph{without parameter updates}, achieving substantial IoU improvements (from 0.63 to 0.93 on medium-triangle tasks) at test time.


\section{Related Work}

\textbf{Human cognition and spatial reasoning.}
Extensive research in cognitive science demonstrates that humans excel at spatial reasoning through processes involving mental rotation, iterative refinement, and visual feedback~\citep{shepard1971mental}. Human spatial cognition operates on fundamental primitives: position, orientation, and scale~\citep{biederman1987recognition,spelke2007core}. Tangram puzzles have been extensively studied as a probe of human spatial intelligence~\citep{bohning1997using}, with developmental work on decomposition and iterative correction~\citep{lee2009enhancing,antrilli2019tangrams,liben2007education}. Recent work on incorporating human cognitive models into AI systems includes approaches that model human mental states~\citep{goodman2016pragmatic}, incorporate human teaching and correction into learning processes~\citep{guo2021learning}, and use human cognitive patterns to make AI systems more interpretable~\citep{lake2017building}. Our framework is inspired by these cognitive mechanisms and explicitly models iterative refinement and feedback-driven correction in continuous geometric space.

\textbf{Test-time self-improvement and refinement.}
Recent work on self-improving AI has explored various mechanisms for enhancing model performance without retraining, including test-time adaptation~\citep{sun2020test}, in-context learning with feedback loops~\citep{madaan2023self}, and reward-guided refinement. 
Methods that leverage iterative refinement, such as self-consistency decoding~\citep{wang2022self} and verifier-guided generation~\citep{cobbe2021training}, demonstrate that models can improve their outputs through multiple rounds of generation and verification. 
In the context of reasoning tasks, approaches like ReAct~\citep{yao2022react} and Reflexion~\citep{shinn2023reflexion} show that language models can refine their reasoning traces based on feedback, while methods in computer vision adapt representations at inference time. 
Our work extends this paradigm to \emph{continuous geometric reasoning}, where we apply reward-based feedback (IoU) to iteratively refine spatial predictions through verifier--refiner loops at test time. 
Unlike prior work that focuses on discrete or symbolic reasoning, we demonstrate test-time self-improvement in a continuous metric space, where \textbf{our geometry verifier serves as instrumentation} to measure geometric consistency and provide feedback for iterative refinement.

\textbf{Evaluation and benchmarks for self-improving systems.}
Evaluation frameworks play a crucial role in assessing self-improvement capabilities. 
Prior work has examined spatial reasoning in multimodal systems through tasks such as mental rotation~\citep{shepard1971mental}, spatial relation matching~\citep{johnson2017clevr}, and compositionality~\citep{wu2023role,hesham2025exploiting}.
Large-scale evaluations report degradation in long-context or complex spatial settings~\citep{ma2024mmlongbench,stogiannidis2025mind}, and prior work on spatial reasoning in multimodal systems has documented systematic limitations in complex spatial settings.
Benchmarks like Winoground~\citep{thrush2022winoground}, \emph{Unfolding Spatial Cognition}~\citep{li2025unfolding}, and \emph{SpatialVLM}~\citep{chen2024spatialvlm} probe spatial reasoning capabilities, while RoboSpatial~\citep{song2025robospatial} and GIQ~\citep{michalkiewicz2025giq} evaluate geometric understanding for robotics and 3D reasoning; specialized models like SpatialVLM still struggle with continuous geometric tasks requiring metric precision.
However, these protocols largely reduce spatial reasoning to symbolic or discrete judgments (e.g., multiple choice, caption matching), without measuring \emph{continuous} geometric errors such as rotation angle drift or sub-pixel position offsets.
In contrast, we introduce a \emph{continuity-space evaluation benchmark} that explicitly measures self-improvement in continuous geometric reasoning, providing a testbed for evaluating refinement mechanisms.

\textbf{Continuous geometric reasoning applications.}
Tangram is a classical tool for studying spatial visualization and compositional reasoning~\citep{bohning1997using}, with computational assembly being NP-hard~\citep{yamada2017comparative,yamada2019solving}.
Recent AI work employs Tangram for abstract visual reasoning~\citep{ji2022abstract,zhao2025master,lin2023compositional,li2026thinking}. 
Li et al.~\citep{li2026thinking} formulate Tangram assembly as a video generation task, but pixel-based generation suffers from geometric inconsistencies. 
Unlike prior efforts emphasizing symbolic inference or implicit visual dynamics, our evaluation operates in \emph{continuity space} with explicit coordinate-based optimization, explicitly quantifying errors in position, angle, and size.
Modern VLMs such as Qwen~\citep{bai2023qwen}, GPT-4o~\citep{islam2025gpt}, and LLaMA variants~\citep{gao2023llama} show impressive language--vision capabilities, yet fundamental geometric invariances (rotation, scale) remain under-tested, with models exhibiting brittle behavior under geometric transformations~\citep{chen2024spatialvlm,michalkiewicz2025giq,huang2025vision}.
Our continuity-space evaluation tailored to Tangram shapes (i) isolates the estimation of position, angle, and size with geometry-aware metrics, and (ii) escalates to two-piece arrangement where error accumulation becomes salient, motivating robust spatial competence for embodied and robotic applications~\citep{song2025robospatial,duan2022survey}.
\section{Methodology}

Each sample is annotated in JSON format with fields \texttt{type}, \texttt{pos}$=[x,y]$, \texttt{angle} (in degrees), and \texttt{size} ($s>0$). 
All shapes are rendered from canonical templates rather than segmented images, eliminating boundary noise and enabling exact polygon-level IoU computation.

\begin{algorithm}[tb]
\caption{Tangram Dataset Pipeline (SVG $\rightarrow$ JSON $\rightarrow$ PNG)}
\label{alg:dataset}
\textbf{Input}: Directory of SVG files (\texttt{IN\_SVG\_DIR}) \\
\textbf{Output}: JSON annotations and optional rendered PNGs
\begin{algorithmic}[1]
\ForAll{ \texttt{svg\_path} $\in$ \texttt{IN\_SVG\_DIR} }
  \State Parse SVG into polygon list $(\texttt{polys}, W, H)$
  \State Fit polygons to canonical tangram templates
  \State Save piece parameters as JSON (\texttt{pos}, \texttt{angle}, \texttt{flip}, \texttt{scale})
  \If{rendering enabled}
     \State Render shapes via geometry engine and save PNG
  \EndIf
  \If{aligned outline available}
     \State Compute IoU between rendered union and outline
  \EndIf
\EndFor
\State \textbf{return} dataset (\texttt{JSON}, \texttt{PNG}, optional IoU logs)
\end{algorithmic}
\end{algorithm}

\begin{algorithm}[t]
\caption{Evaluation Pipeline}
\label{alg:eval-pipeline}
\textbf{Input:} \texttt{IN\_DIR} (PNGs), \texttt{GT\_DIR} (JSONs), \texttt{OUT\_DIR}, model, mode \\
\textbf{Output:} Pred JSONs, metrics (L2/angle/size/IoU), visualizations
\begin{algorithmic}[1]
\ForAll{$I \in \texttt{IN\_DIR}$}
  \State $G \leftarrow$ pair JSON; $\texttt{two}\leftarrow$ \textsc{IsTwoPiece}$(G)$
  \State Predict: call model; parse \& validate by \texttt{mode}
  \State Compute metrics (if $G$ exists): L2/angle/size/IoU
  \State Render GT/PRED/OVERLAY; save outputs
\EndFor
\end{algorithmic}
\end{algorithm}

We design four tasks that progressively increase geometric difficulty:
\begin{enumerate}
    \item \textbf{pos-only}: fix GT $(\alpha,s)$, predict $\hat{\mathbf{p}}$;
    \item \textbf{angle-only}: fix GT $(\mathbf{p},s)$, predict $\hat{\alpha}$;
    \item \textbf{size-only}: fix GT $(\mathbf{p},\alpha)$, predict $\hat{s}$;
    \item \textbf{two-piece arrangement}: fix each piece’s $(\alpha,s)$, predict two positions $(\hat{\mathbf{p}}_1,\hat{\mathbf{p}}_2)$, and assess composition.
\end{enumerate}
We also include a \textbf{joint} setting that predicts all three fields simultaneously to expose results.

\subsection{Metrics and Inference Protocol}
We report the rasterized intersection-over-union (IoU) between predicted and ground-truth shapes 
on a $512\times512$ canvas, using 1–2\,px dilation for morphology-tolerant evaluation. 
For two-piece assembly, IoU is computed over the union of both pieces to capture overlap penalties.

We evaluate four Vision--Language Models (VLMs): Qwen-3B, Qwen-72B, GPT-4o mini, and LLaMA Maverick. 
Each model receives a Tangram silhouette image and is prompted to output a minimal JSON containing the requested field(s): position (\texttt{pos}), orientation (\texttt{angle}), or scale (\texttt{size}). 
All predictions are evaluated in a normalized $[0,1]^2$ coordinate frame. 
Geometric consistency is assessed using \texttt{geometry}, which computes Euclidean position error, angular deviation, and scale difference, 
and \texttt{overlay}, which renders predicted and ground-truth polygons to measure intersection-over-union (IoU). 
We report results under both zero-shot and few-shot ICL settings (typically $k=15$), following a unified inference and evaluation protocol across all models.

\begin{figure*}[t]
  \centering
  \includegraphics[width=0.85\textwidth]{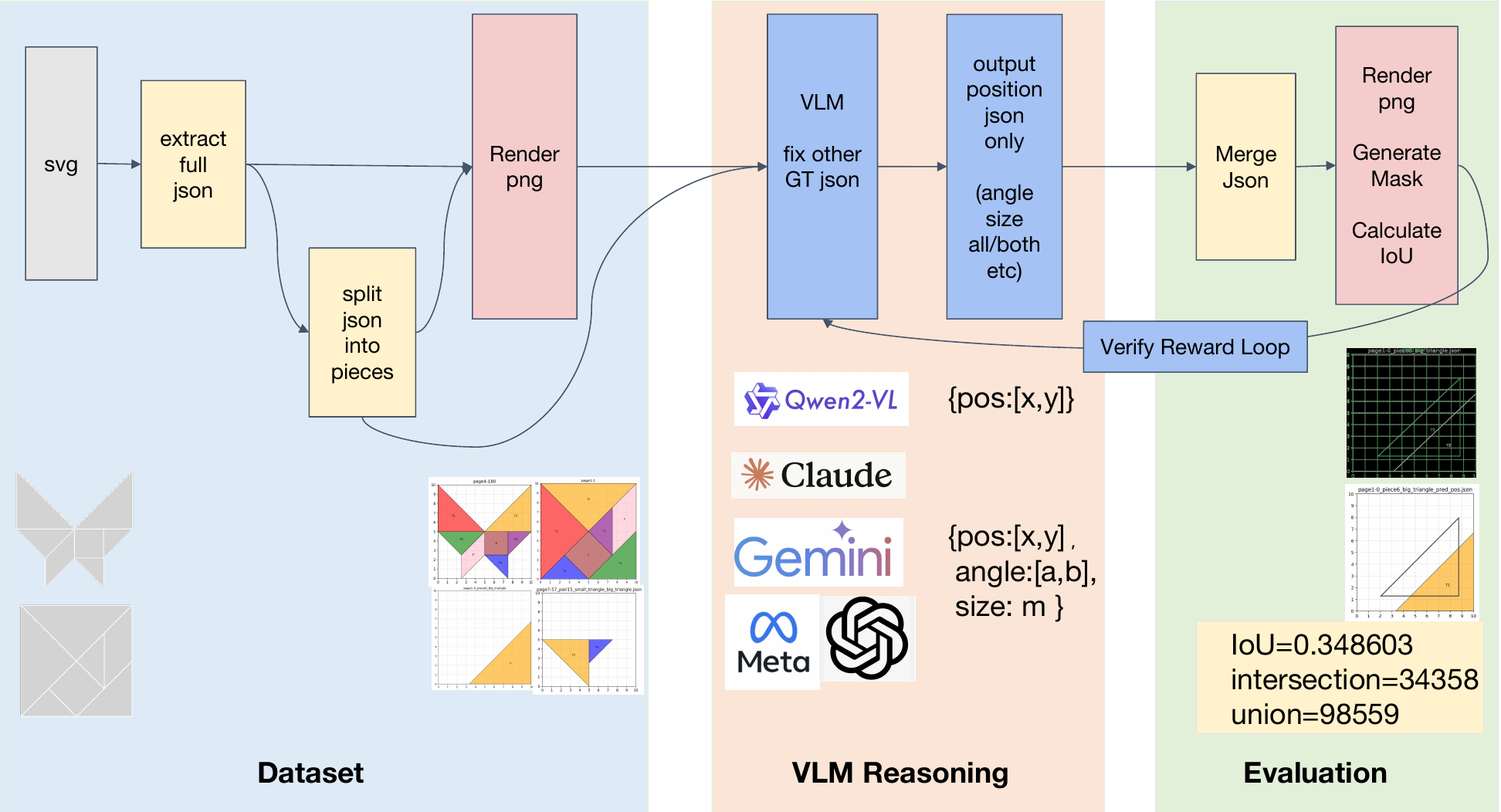}
  \caption{Overall dataset construction pipeline from SVG $\rightarrow$ JSON $\rightarrow$ PNG $\rightarrow$ split tasks. 
  The diagram shows how raw SVG tangram silhouettes are parsed into JSON annotations (type, position, angle, size), 
  rendered into training/evaluation images, and split into single-piece, two-piece, or full-tangram subsets.}
  \label{fig:dataset-pipeline}
\end{figure*}
\subsection{Dataset and Tasks (Continuity-Space Protocol)}
\label{sec:dataset-tasks}
We build a Tangram benchmark with two splits:
\textbf{Single-piece}: each image contains a \emph{single} canonical piece placed with GT $(\mathbf{p},\alpha,s)$. This supports pos/angle/size decoupling.
\textbf{Two-piece}: each image contains two pieces whose relative arrangement matters (mutual non-overlap coverage).

\subsection{Test-Time Self-Improvement via Reward-Guided Refinement}
\label{sec:icl-loop}

\textbf{Setting.}
For a single tangram piece, the pose is
$\Theta=(\mathbf{p},\alpha,s)$ with position $\mathbf{p}\in[0,10]^2$,
angle $\alpha$ (deg), and size $s>0$.
Let $\mathcal{U}(\Theta)$ be the rendered polygon from the canonical
template under $(\mathbf{p},\alpha,s)$, and $S$ the ground-truth polygon.

\paragraph{Reward (what we actually optimize).}
We use a scalar reward that trades off geometric coverage (IoU) against
position error only:
\begin{align}
\mathcal{R}(\Theta)
\;=\;
\mathrm{IoU}\!\left(\mathcal{U}(\Theta),\, S\right)
\;-\;
\lambda \cdot \frac{\lVert \hat{\mathbf{p}}-\mathbf{p}\rVert_2}{10}.
\label{eq:reward}
\end{align}
Here $\hat{\mathbf{p}}$ is the GT position, the canvas side length is $10$
(hence the division by $10$ for normalization), and $\lambda>0$ is a small
weight. \emph{No} overlap penalty, edge-shape term, angle/scale regularizer,
or global loss is used in our implementation.

\paragraph{Self-refinement loop mechanics (training-free).}
We run $T$ iterations of self-refinement through ICL + feedback:
(i) build $k$ few-shot pairs \emph{excluding} the current sample;
(ii) query the VLM with a minimal JSON instruction for the desired field(s),
and from the second iteration onward append a numeric feedback hint of the form
``\texttt{previous IoU=\$x.xx. Try a small correction }$(\Delta x,\Delta y)$\texttt{.}'';
(iii) keep the candidate with the highest $\mathcal{R}$ in Eq.~\eqref{eq:reward},
with early stop once $\mathrm{IoU}\ge\tau$.

\paragraph{Deterministic local refinement.}
If $\mathrm{IoU}<\tau$ and the task involves position (\texttt{pos} or \texttt{all}),
we perform a tiny grid search around the current best $(x,y)$ using a
$3\times3$ neighborhood at step sizes $0.6\!\rightarrow\!0.3\!\rightarrow\!0.15$
(canvas units). We accept the first move that increases IoU and update the
best candidate; finally we recompute the reward
$\mathcal{R}=\mathrm{IoU}-\lambda\cdot(\mathrm{L2}/10)$.
This self-refinement loop is \emph{training-free}, uses only CPU, and treats the VLM as a
proposal generator with a geometry-based verifier that guides iterative improvement at test time.

\begin{algorithm}[t]
\caption{VLM + ICL + Reward Loop (simplified)}
\label{alg:vlm-icl-loop}
\textbf{Input:} image $I$, model $M$, mode $\in\{\text{pos},\text{angle},\text{size},\text{all}\}$, 
ICL size $k$, loop iters $T$, threshold $\tau$ \\
\textbf{Output:} best JSON prediction $J^\star$, best IoU
\begin{algorithmic}[1]
\State $\mathcal{S} \leftarrow$ sample $k$ few-shot (image, JSON) pairs for ICL
\State Initialize $\text{best} \leftarrow (\text{iou}=0,\ J=\varnothing)$
\For{$t=1$ to $T$}
  \State Query $M$ with $I$ + $\mathcal{S}$ + refinement hint
  \State Parse output $\rightarrow J_t$ (JSON fields)
  \State Compute $\text{iou}_t = \textsc{IoU}(J_t,\ G)$
  \If{$\text{iou}_t > \text{best.iou}$} $\text{best}\leftarrow(J_t,\ \text{iou}_t)$ \EndIf
  \If{$\text{best.iou} \ge \tau$} \textbf{break} \EndIf
\EndFor
\State Optionally run small local search around $\text{best.pos}$
\State \textbf{return} $J^\star = \text{best.J},\ \text{best.iou}$
\end{algorithmic}
\end{algorithm}

\begin{figure}[t]
  \centering
  \begin{subfigure}[b]{0.48\columnwidth}
    \centering
    \includegraphics[width=\textwidth]{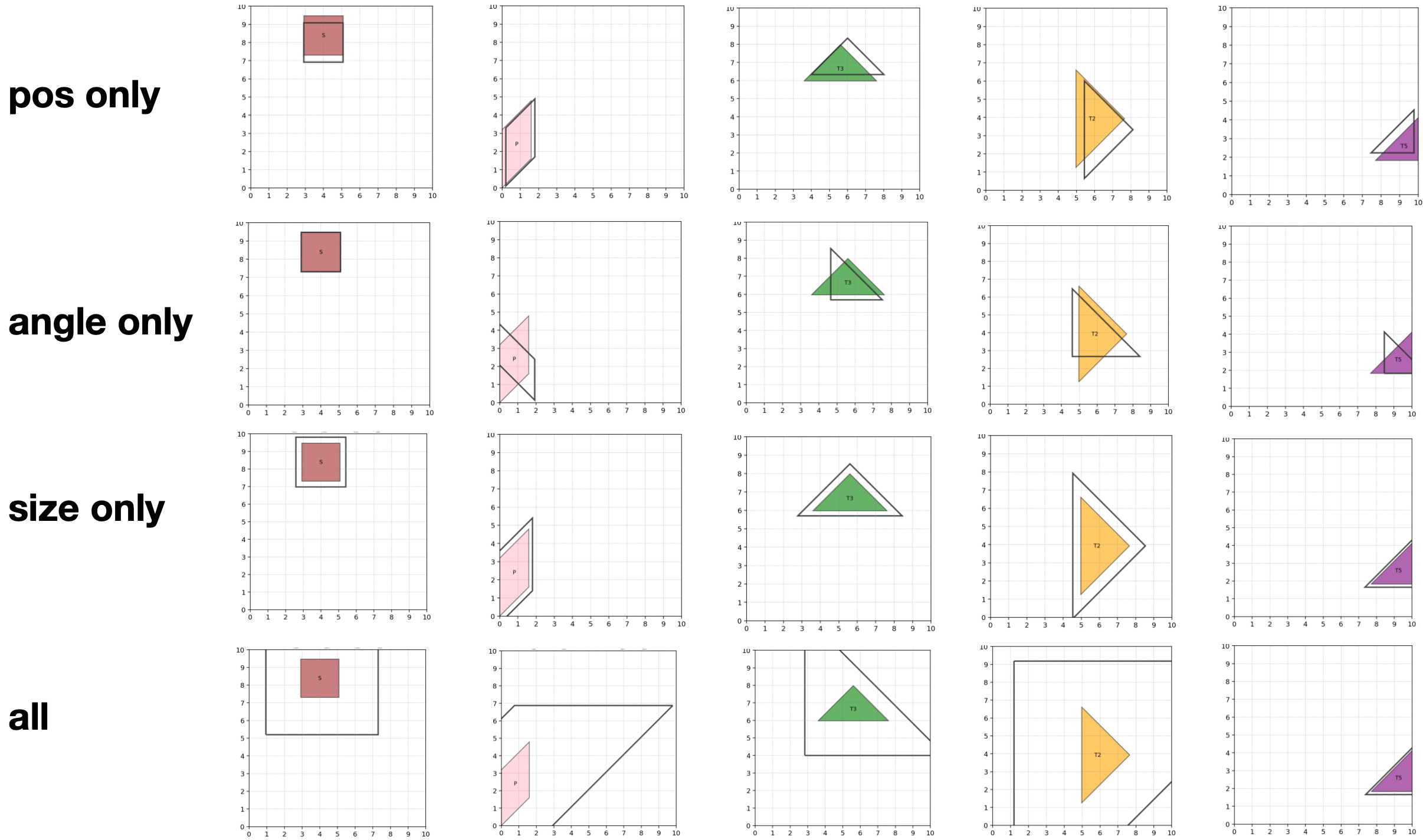}
    \caption{Single-piece setup and outputs: position, angle, size, and all.}
    \label{fig:one-piece}
  \end{subfigure}
  \hfill
  \begin{subfigure}[b]{0.48\columnwidth}
    \centering
    \includegraphics[width=0.9\textwidth]{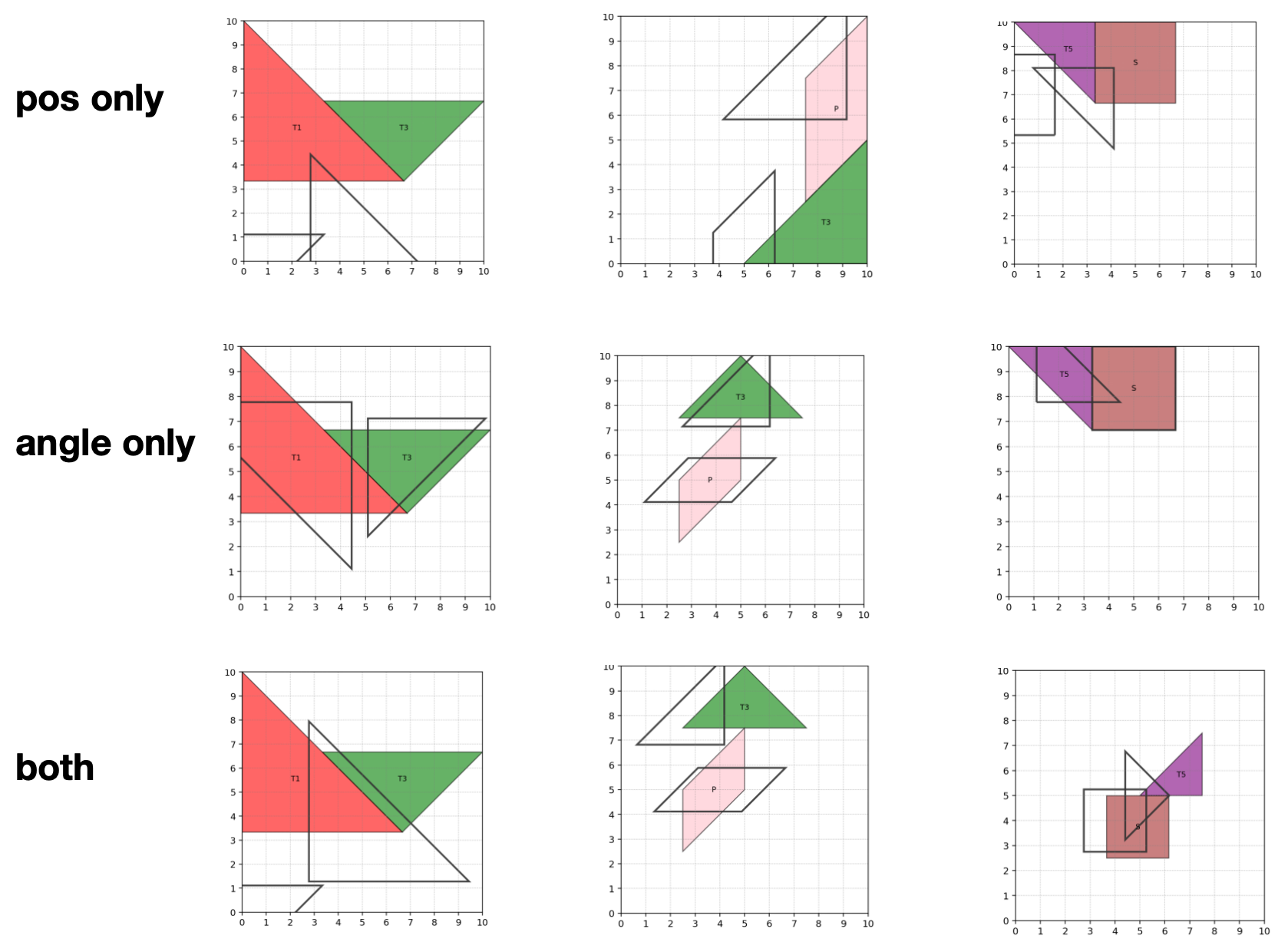}
    \caption{Two-Piece setup and outputs: position, angle and both.}
    \label{fig:two-piece}
  \end{subfigure}
  \caption{Spatial reasoning tasks: single-piece and two-piece Tangram assembly.}
  \label{fig:spatial-tasks}
\end{figure}

\section{Results and Analysis}

\subsection{Part I: Cross-Model Comparison on Spatial Reasoning}
\label{sec:vlm-spatial}
\textbf{Setup.} 
We evaluate \emph{pos-only}, \emph{angle-only}, \emph{size-only}, and \emph{joint (all)} predictions across five models: Qwen-3B, Qwen-72B, GPT-4o mini, LLaMA Maverick, and Gemini-2.5-pro. 
We report mean$\pm$95\%CI over the test set and include equivariance stress-tests (rotation, mirror, and scale). 
Unlike prior sections, we unify cross-model and factorized testing into a single comparison table, where each row is a model and each column corresponds to a prediction task. 
Metrics reported are task-specific errors (L2, angular degrees, relative scale) and IoU (higher is better).

\textbf{Findings.} 
(i) Larger models reduce L2 and scale errors, but angle remains fragile across all models; 
(ii) joint prediction aggregates noise across multiple axes, amplifying errors versus factorized tasks; 
(iii) IoU is highly sensitive to angular mismatch even when position errors are small.

\begin{table*}[htbp]
  \centering
  \small
  \setlength{\tabcolsep}{6pt}
  \caption{Unified results for VLM one-piece spatial reasoning (\(\uparrow\) higher is better). 
VLM performance is relatively low compared to human performance~\citep{bohning1997using}.}
  \label{tab:onepiece}  
  \begin{tabular*}{\textwidth}{@{\extracolsep{\fill}}lcccc}
    \toprule
    \textbf{Method} & \textbf{Pos IoU $\uparrow$} & 
    \textbf{Angle IoU $\uparrow$} & 
    \textbf{Size IoU $\uparrow$} & 
    \textbf{All IoU $\uparrow$} \\
    \midrule
    Claude-Sonnet-4 & 0.419 & 0.394 & 0.372 & 0.395 \\
    Gemini-2.5-pro & \textbf{0.443} & \textbf{0.434} & \textbf{0.432} & \textbf{0.417} \\
    GPT-4o mini-8B & 0.427 & 0.429 & 0.393 & 0.413 \\
    LLaMA Maverick 17B & 0.424 & 0.427 & 0.371 & 0.377 \\
    Qwen-3B & 0.236 & 0.414 & 0.369 & 0.219 \\
    Qwen-72B & 0.415 & 0.432 & 0.425 & 0.408 \\
    \bottomrule
  \end{tabular*}
\end{table*}

\subsection{Part II: Spatial Arrangement (Two-Piece Composition)}
\label{sec:part3}
\textbf{Setup.} We test \emph{arrangement} with two pieces. We consider three modes: (A) fix both $(\alpha,s)$, predict $(\mathbf{p}_1,\mathbf{p}_2)$; (B) fix $\mathbf{p},s$ and predict \emph{angles}; (C) predict positions + angles jointly (scaled fixed). Metrics: union IoU and overlap penalty.

\textbf{Findings.} Arrangement is \emph{significantly} harder than single-piece: IoU drops $\sim$0.3 even when single-piece IoU exceeds 0.7. Typical failure modes: mutual collision, near-miss adjacency, and mirrored angles producing plausible shapes but wrong unions, as shown on figure 1.


\begin{table}[t]
  \centering
  \footnotesize
  \setlength{\tabcolsep}{4pt} 
 \caption{Unified results for VLM two-piece spatial reasoning (\(\uparrow\) higher is better). 
VLM performance is relatively low compared to human performance~\citep{bohning1997using}.}
  \label{tab:vlm-spatial}
  \begin{tabular}{lccc}
    \toprule
    \textbf{Model} & \textbf{Pos IoU \(\uparrow\)} & \textbf{Angle IoU \(\uparrow\)} & \textbf{All IoU \(\uparrow\)} \\
    \midrule
    Claude-Sonnet-4 & 0.318 & 0.394 & 0.235 \\
    Gemini-2.5-pro  & \textbf{0.340} & 0.397 & 0.340 \\
    GPT-4o mini     & 0.276 & 0.394 & 0.278 \\
    LLaMA Maverick  & 0.220 & 0.427 & \textbf{0.371} \\
    Qwen-3B         & 0.192 & 0.317 & 0.214 \\
    Qwen-72B        & 0.253 & \textbf{0.495} & 0.248 \\
    \bottomrule
  \end{tabular}
\end{table}

\subsection{Part III: Test-Time Self-Improvement via Reward-Guided Refinement}
\label{sec:part4}
\textbf{Setup.} We focus on the \emph{medium triangle} subset (single-piece), start from the VLM's JSON output, run $T$ self-refinement loop iterations with reward $\mathcal{R}$. We allow a tiny local search over $\mathbf{p}$ at the end if IoU remains low.

\begin{table*}[ht]
\centering
\small
\caption{Medium triangle IoU across different settings (baseline start = 0.65).}
\label{tab:loop}
\begin{tabular}{ccccccc}
\toprule
\textbf{Setting Number} & \textbf{Description} & \textbf{ICL (k)} & \textbf{Loop} & \textbf{Threshold} & \textbf{Temp.} & \textbf{IoU (final)} \\
\midrule
1 & \texttt{VLM + ICL + Loop \phantom{ICL + }}                 & 15  & 6   & 0.9 & 0   & \textbf{0.9320} \\
2 & \texttt{VLM + \phantom{ICL + }Loop \phantom{ICL + }}       & n/a & 6   & 0.9 & 0   & 0.9320 \\
3 & \texttt{VLM + ICL + Loop \phantom{ICL + }}                 & 20  & 6   & 0.9 & 0   & 0.9300 \\
4 & \texttt{VLM + ICL\phantom{ + Loop} \phantom{ICL + }}       & 15  & n/a & n/a & 0   & 0.7950 \\
5 & \texttt{VLM + ICL  \phantom{ICL + } + temp} & 15  & n/a & n/a & 0.5 & 0.7690 \\
6 & \texttt{VLM only\phantom{ + tLoop} \phantom{ICL + }}  & n/a & n/a & n/a & 0   & 0.6500 \\
\bottomrule
\end{tabular}
\end{table*}

\begin{figure}[t]
  \centering
  \includegraphics[width=0.8\columnwidth]{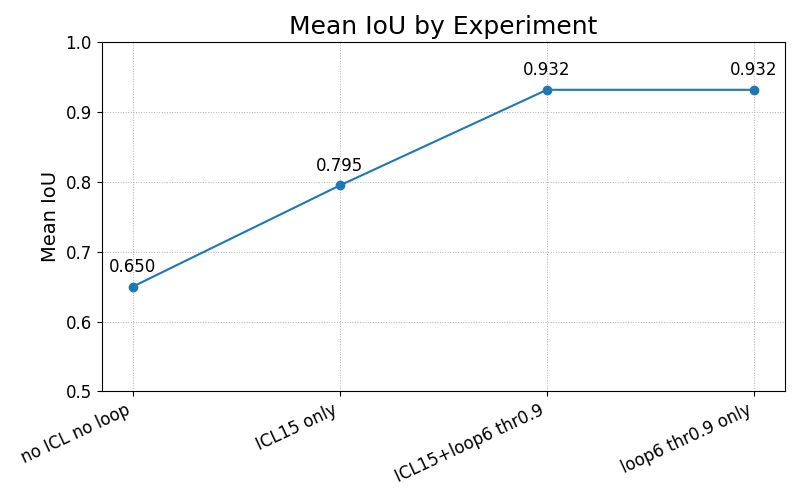}
  \caption{Mean IoU across ablations on the \emph{medium triangle}. 
  The test-time self-refinement loop (ICL + reward) yields the largest gain.}
  \label{fig:icl-loop-iou}
\end{figure}

\begin{table}[ht]
\centering
\small
\caption{Ablation on loop count and threshold (keep ICL = 15 fixed). Baseline IoU = 0.65.}
\label{tab:loop-ablation}
\begin{tabular}{ccccc}
\toprule
\textbf{Setting Number} & \textbf{Description} & \textbf{Loop} & \textbf{Threshold} & \textbf{IoU (final)} \\
\midrule

1  & ICL + Loop & 6 & 0.9 & \textbf{0.9320} \\
7  & ICL + Loop & 4 & 0.9 & 0.9287 \\
8  & ICL + Loop & 2 & 0.9 & 0.9291 \\
\midrule 
9 & ICL + Loop & 6 & 0.5 & 0.8410 \\
10 & ICL + Loop & 4 & 0.5 & 0.8609 \\
11 & ICL + Loop & 2 & 0.5 & 0.7200 \\
\midrule 
12  & ICL + Loop & 6 & 0.8 & 0.9310 \\
13 & ICL + Loop & 6 & 0.7 & 0.9233 \\
14 & ICL + Loop & 6 & 0.6 & 0.9063 \\
\midrule 
15  & ICL + Loop & 8 & 0.9 & 0.9345 \\
16  & ICL + Loop & 10 & 0.9 &  0.9258\\
17  & ICL + Loop & 12 & 0.9 & 0.9323 \\
\bottomrule
\end{tabular}
\end{table}

\begin{table}[ht]
\centering
\small
\caption{Ablation on ICL window size ($k$), keep loop and threshold constant.}
\label{tab:icl-loop}
\begin{tabular}{cccc}
\toprule
\textbf{ICL ($k$)} & \textbf{Loop} & \textbf{Threshold} & \textbf{IoU (final)} \\
\midrule
15 & 8  & 0.90 & 0.9345 \\
20 & 8  & 0.90 & 0.9311 \\
25 & 8  & 0.90 & 0.9310 \\
\bottomrule
\end{tabular}
\end{table}

\paragraph{Findings.}
Table~\ref{tab:loop} compares six configurations (Settings~1--6), illustrating how ICL, loop refinement, and temperature jointly affect performance. Figure~\ref{fig:icl-loop-iou} visualizes mean IoU across these ablations on the medium triangle.

\begin{itemize}
    \item \textbf{VLM only vs. ICL (Setting 6 vs.~Setting 4).} 
    The plain VLM baseline (0.65) improves to 0.795 with \textit{ICL15}, yielding an absolute gain of \textbf{+0.145}. 
    However, raising the temperature to 0.5 (Setting~5) slightly degrades IoU to 0.769 (\textbf{--0.026}), 
    suggesting that while ICL stabilizes generation, excessive sampling diversity introduces numerical noise that harms geometric consistency.

    \item \textbf{VLM only vs. Loop only (Setting 6 vs.~Setting 2).} 
    Running the geometry-based \textbf{Loop} (6 iterations, threshold 0.9) elevates IoU from 0.65 to \textbf{0.932} (\textbf{+0.282}), 
    far exceeding the ICL-only gain. 
    This demonstrates that the \emph{test-time self-refinement loop} effectively corrects small positional and angular errors that token-based generation cannot fix.

    \item \textbf{ICL + Loop comparison (Settings~1--3, 15--17).}

Combining ICL with the loop (Setting~1, 0.932) achieves nearly the same performance as Loop only (Setting~2, 0.932), 
while a larger ICL window ($k{=}20$, Setting~3) yields slightly lower IoU (0.930). 
When extending the loop length further (Settings~15--17, with 8–12 iterations), 
we observe that IoU improvements become marginal: 0.9345, 0.9258, and 0.9323 respectively. 
This indicates that the refinement process quickly saturates, as most geometric errors are already corrected within six iterations. 
In our analysis, we consider an IoU increase above 0.01 as statistically significant; 
thus, we treat 0.932 as the effective performance ceiling under the current framework.

\item \textbf{General trend.} 
Comparing Settings~1--6 collectively, the \textbf{Loop mechanism contributes the majority of improvement}, 
while ICL provides consistent but smaller gains. 
Overly high temperature or large ICL windows introduce instability, confirming that precision feedback, not randomness, is key to continuous geometric refinement. 
Furthermore, as shown in Table~\ref{tab:icl-loop}, when the loop count and threshold are held constant, 
increasing the ICL window size ($k$) from 15 to 25 does not yield further improvement; 
instead, IoU slightly decreases from 0.9345 to 0.9310. 
This suggests that excessively large ICL contexts may introduce redundancy or noise, 
and that moderate exemplars (\(k{=}15\)) strike a more effective balance between stability and precision.

\end{itemize}

\noindent\textbf{Takeaways.}
(1) The \textbf{self-refinement loop} contributes more than ICL overall, acting as a geometry-aware \emph{test-time improvement mechanism} correcting small continuous errors;  
(2) \textbf{ICL+Loop} reaches the same ceiling as \textit{Loop only}, while improving initialization and convergence stability;  
(3) \textbf{Excessive ICL} (larger $k$ or higher temperature) introduces noise, causing slight degradation;  
(4) The recommended setting is \textbf{ICL$\,k{=}15$ + Loop$\,{=}6$ + $\tau{=}0.9$ + temperature$\,{=}0$}, balancing stability and accuracy for effective test-time self-improvement.  

\paragraph{Loop Parameter Sensitivity (Settings 7–14).}
Table~\ref{tab:loop-ablation} compares different loop counts and acceptance thresholds under constant ICL ($k{=}15$).  
By contrasting Settings~8 (2 loops, 0.9291), 7 (4 loops, 0.9287), and 1 (6 loops, 0.9320), we observe that performance quickly saturates after only a few refinement iterations, indicating that most geometric errors are corrected early in the process.  
Varying the acceptance threshold further highlights its decisive role in stability: comparing Settings~1 ($\tau{=}0.9$, 0.9320) with 9 ($\tau{=}0.5$, 0.8410) and 11 ($\tau{=}0.5$, 0.7200), a steep IoU decline appears as the gating becomes more lenient.  
This degradation shows that low thresholds admit noisy or suboptimal corrections, breaking geometric consistency, whereas higher thresholds act as a filter that stabilizes updates.  
Across Settings~12–14 (6 loops, $\tau$ decreasing from 0.8 to 0.6), the same trend persists: tightening the gate consistently improves stability while additional iterations beyond six offer diminishing returns.  
Overall, these results indicate that accuracy depends mainly on the threshold choice, while increasing loop depth yields limited additional benefits.


\section{Discussion and Limitations}
\label{sec:discussion}

\textbf{Results summary.}
Our evaluation across single-piece and two-piece Tangram assembly reveals that most VLMs achieve moderate IoU (0.2--0.45) on single-piece tasks, dropping to 0.23 on two-piece composition due to error compounding. 
\textbf{Recursive self-refinement} consistently improves alignment, with loops converging within 1--2 iterations and achieving IoU gains of over 40\% without parameter updates. 
The refinement is \textbf{recursive} because each iteration updates the ICL context and feedback signals based on previous outputs, forming a closed-loop process. 
Our reward-guided framework generalizes beyond Tangram to any task requiring continuous spatial alignment, including robotics manipulation and navigation~\citep{song2025robospatial,duan2022survey}, where geometric precision is critical.

Despite strong performance on semantic benchmarks, VLMs perform poorly in continuous-space evaluation when operating in single-pass mode, achieving only $\approx$0.41 IoU compared to human performance where children can complete Tangram tasks successfully~\citep{bohning1997using}. 
We attribute this to three factors: (1) \textbf{Training distribution mismatch}: VLMs are trained on discrete semantic tasks, rarely encountering precise coordinate prediction, unlike humans who develop spatial reasoning through experience with continuous geometric tasks; 
(2) \textbf{Output mismatch}: autoregressive decoders force continuous quantities into discrete tokens, causing rounding and numerical drift, whereas humans reason directly in continuous space; 
(3) \textbf{Absence of geometry-aware feedback}: models lack geometric consistency supervision during training, while humans naturally receive visual feedback that guides spatial refinement.

\textbf{Limitations and safety considerations.} 
Our evaluation is limited to one- and two-piece configurations rather than full seven-piece assembly. 
IoU computation is rasterization-based and resolution-sensitive (512×512, validated at 1024×1024). 
The reward function uses a scalar $\lambda$, and our benchmark focuses on canonical shapes; future work may explore adaptive schedules and irregular or noisy silhouettes. 
Regarding \textbf{stability and safety}: refinement loops converge reliably within 1--2 iterations for small errors, but may exhibit regression risk if initial predictions are severely misaligned. 
In robotics applications, this requires careful validation of convergence and fallback mechanisms to prevent unsafe behaviors.

\textbf{Connection to long-term self-improvement and evolutionary methods.}
While our framework focuses on test-time refinement, it can be extended to long-term self-improvement by accumulating successful strategies across episodes and adapting reward schedules based on performance. 
This connects to evolutionary approaches where successful patterns are preserved and refined over time, though operating in continuous rather than discrete space.


\section{Conclusion and Future Work}
We presented a \textbf{test-time self-improvement framework} for vision--language models (VLMs) operating in continuous geometric space. 
By introducing reward-guided refinement loops, we demonstrate that VLMs can substantially enhance their geometric reasoning capabilities \emph{without retraining}, achieving significant IoU improvements and moving self-improving AI from promise to practice in continuous spatial domains. 
Our continuous-space evaluation benchmark explicitly measures spatial reasoning through geometric metrics, revealing systematic weaknesses in current VLMs and providing a testbed for evaluating self-improvement mechanisms. 
Looking forward, we plan to extend our framework to more complex geometric reasoning tasks, explore adaptive reward schedules, investigate stability and alignment properties of refinement loops, and conduct human with VLM comparison studies. 
Our work contributes to the broader vision of self-improving AI systems that can adapt and enhance their capabilities through feedback-driven refinement, particularly in domains where continuous precision matters. 
While current VLMs remain far below human-level spatial reasoning capabilities, our recursive refinement framework demonstrates that test-time self-improvement can substantially narrow this gap, moving toward human-inspired reasoning processes that incorporate iterative feedback and geometric consistency.

\bibliography{iclr2026_conference}
\bibliographystyle{iclr2026_conference}

\appendix
\section{Appendix}

\subsection{The Use of LLMs}
We use llm to check and correct grammar and spelling mistakes. In addition, we also use llm to polish the sentences in our paper to make them more fluent.

\end{document}